\documentclass[lettersize,journal]{IEEEtran}

\usepackage{caption}
\usepackage{subcaption}
\usepackage{amsmath,amsfonts}
\usepackage{array}
\usepackage{textcomp}
\usepackage{stfloats}

\usepackage{url}
\usepackage{verbatim}
\usepackage{graphicx}
\usepackage{cite}
\usepackage{multirow}
\usepackage{booktabs}

\usepackage{xcolor}

\usepackage{xspace}
\newcommand{\name}[0]{NTFormer\xspace}
\newcommand{\tname}[0]{Node2Par\xspace}

\usepackage{bbding}
\makeatletter
\DeclareRobustCommand\onedot{\futurelet\@let@token\@onedot}
\def\@onedot{\ifx\@let@token.\else.\null\fi\xspace}
\def\eg{\emph{e.g}\onedot} 
\def\ie{\emph{i.e}\onedot}

\def\etal{\emph{et al}\onedot}

\usepackage{scalerel}
\usepackage{tikz}
\usetikzlibrary{svg.path}
\definecolor{orcidlogocol}{HTML}{A6CE39}
\tikzset{
    orcidlogo/.pic={
        \fill[orcidlogocol] svg{M256,128c0,70.7-57.3,128-128,128C57.3,256,0,198.7,0,128C0,57.3,57.3,0,128,0C198.7,0,256,57.3,256,128z};
        \fill[white] svg{M86.3,186.2H70.9V79.1h15.4v48.4V186.2z}
        svg{M108.9,79.1h41.6c39.6,0,57,28.3,57,53.6c0,27.5-21.5,53.6-56.8,53.6h-41.8V79.1z M124.3,172.4h24.5c34.9,0,42.9-26.5,42.9-39.7c0-21.5-13.7-39.7-43.7-39.7h-23.7V172.4z}
        svg{M88.7,56.8c0,5.5-4.5,10.1-10.1,10.1c-5.6,0-10.1-4.6-10.1-10.1c0-5.6,4.5-10.1,10.1-10.1C84.2,46.7,88.7,51.3,88.7,56.8z};
    }
}
\newcommand\orcidicon[1]{\href{https://orcid.org/#1}{\mbox{\scalerel*{
                \begin{tikzpicture}[yscale=-1,transform shape]
                \pic{orcidlogo};
                \end{tikzpicture}
            }{|}}}}
\usepackage{hyperref}
\hypersetup{hidelinks,
	colorlinks=true,
	allcolors=black,
	pdfstartview=Fit,
	breaklinks=true}

\usepackage{algorithmicx}
\usepackage{algpseudocode}  
\usepackage{algorithm}  

\begin{document}

\title{NTFormer: A Composite Node Tokenized Graph Transformer for Node Classification}

\author{Jinsong Chen\textsuperscript{\orcidicon{0000-0001-7588-6713}},  
Siyu Jiang\textsuperscript{\orcidicon{0009-0000-4565-8402}},
Kun He\textsuperscript{\orcidicon{0000-0001-7627-4604}}, \IEEEmembership{Senior~Member,~IEEE}
\IEEEcompsocitemizethanks{\IEEEcompsocthanksitem J. Chen is with School of Computer Science and Technology, Huazhong University of Science and Technology; 
Institute of Artificial Intelligence, Huazhong University of Science and Technology; 
and Hopcroft Center on Computing Science, Huazhong University of Science and Technology, Wuhan 430074,  China. 
\IEEEcompsocthanksitem S. Jiang and K. He are with School of Computer Science and Technology, Huazhong University of Science and Technology; and Hopcroft Center on Computing Science, Huazhong University of Science and Technology, Wuhan 430074,  China. 
}
\thanks{The first two authors contribute equally.}
\thanks{(Corresponding author: Kun He. E-mail: brooklet60@hust.edu.cn.)}
}
\markboth{Journal of \LaTeX\ Class Files,~Vol.~14, No.~8, August~2021}%
{Shell \MakeLowercase{\textit{et al.}}: A Sample Article Using IEEEtran.cls for IEEE Journals}


\maketitle

\begin{abstract}
Recently, the emerging graph Transformers have made significant advancements for node classification on graphs. In most graph Transformers, a crucial step involves transforming the input graph into token sequences as the model input, enabling Transformer to effectively learn the node representations. However, we observe that existing methods only express partial graph information of nodes through single-type token generation. Consequently, they require tailored strategies to encode additional graph-specific features into the Transformer to ensure the quality of node representation learning, limiting the model flexibility to handle diverse graphs. To this end, we propose a new graph Transformer called NTFormer to address this issue. NTFormer introduces a novel token generator called Node2Par, which constructs various token sequences using different token elements for each node. This flexibility allows Node2Par to generate valuable token sequences from different perspectives, ensuring comprehensive expression of rich graph features. Benefiting from the merits of Node2Par, NTFormer only leverages a Transformer-based backbone without graph-specific modifications to learn node representations, eliminating the need for graph-specific modifications. Extensive experiments conducted on various benchmark datasets containing homophily and heterophily graphs with different scales demonstrate the superiority of NTFormer over representative graph Transformers and graph neural networks for node classification.

\end{abstract}

\begin{IEEEkeywords}
Graph Transformer, Token Generator, Node Classification, Node2Par.
\end{IEEEkeywords}

\section{Introduction}

\IEEEPARstart{N}{ode} classification, which aims to predict the labels of nodes on graphs, is one of the most important tasks in the field of graph data mining.
Due to its wide applications in various real-world scenarios such as biology analysis~\cite{biology}, node classification has attracted great attention in recent years.
Graph neural networks (GNNs)~\cite{gcn,gat,gcnii} have become the leading architecture for this task.
Benefiting from the massage-passing mechanism~\cite{mpnn}, GNNs recursively aggregate information of nodes through the adjacency matrix.
By integrating structural property with the node features, the modeling capacity of GNNs is substantially enhanced for learning distinguishable node representations from graphs. 

Despite their remarkable performance, 
recent studies~\cite{oversmoothing,oversq} have revealed the inherent limitations of the message-passing mechanism, \ie over-smoothing~\cite{oversmoothing} and over-squashing~\cite{oversq} issues.
These situations hinder GNNs from deeply capturing complex graph structural information, \eg, long-range dependencies, adversely affecting the model potential for graph mining tasks.

Recently, graph Transformers~\cite{graphormer,gt,gps,magt} have emerged as a promising alternative to GNNs due to the powerful modeling capacity of the Transformer architecture.
The key idea is to utilize the self-attention mechanism of Transformers~\cite{transformer} to learn node representations from complex graph features.
With delicate designs of encoding various graph structural information~\cite{san,graphormer,agt} into the Transformer architecture, graph Transformers have showcased remarkable capabilities for various graph data mining tasks.
In the context of node classification, existing graph Transformers can be broadly categorized into two categories based on the model input: entire graph-based methods and token sequence-based methods.

Entire graph-based graph Transformers~\cite{gt,san,gps,agt,signgt} require the whole graph as the model input.
These methods focus on explicitly transforming the structural information into feature vectors or structural-aware biases, such as eigenvectors of the Laplacian matrix~\cite{gt,san}. 
Then, the Transformer-based backbone extended with extracted structural information is adopted to learn node representations from the input graph.
Instead of inputting the entire graph, token sequence-based graph Transformers~\cite{gophormer,ansgt,nagphormer} construct a series of token sequences for each node, consisting of different types of token elements (\eg node-based~\cite{gophormer,ansgt} and neighborhood-based~\cite{nagphormer,nag+} tokens).
The generated token sequences are further fed to a Transformer-based backbone to learn node representations for downstream machine learning tasks.
Moreover, the entire graph-based graph Transformers could also be regarded as a special variant of token sequence-based methods by treating the input graph as a long token sequence containing all nodes.

Rethinking the purpose of constructing the token sequence, 
the process can be seen as an implicit transformation of complex graph features into tokens, which preserves relevant and necessary graph information for each node to learn informative node representations.
Assuming one would like to describe a node on the graph, the constructed token sequences serve as sentences derived from the graph to express the crucial characteristics of the target node.
In previous studies~\cite{gophormer,ansgt,nagphormer,nag+}, node and neighborhood are two key elements for constructing token sequences, conveying the extracted graph information from different levels.
Node-oriented token sequences can address the issue of over-squashing and capture long-range dependencies but are inefficient in preserving local topological information.
On the other hand, neighborhood-oriented token sequences can accurately reflect topological features but often suffer from the over-squashing problem and fail to express node-level properties on graphs, such as heterophily~\cite{vcrgt}.

The aforementioned phenomenon indicates that single-type oriented token sequences can only express a partial view of the graph information for each node, which hampers the learning of node representations.
Consequently, 
existing methods necessitate tailored designs, such as topology-aware attention biases~\cite{gophormer,ansgt} or node-wise positional encoding~\cite{nagphormer}, to learn distinct node representations and achieve competitive performance.
However, such delicate modifications could potentially limit the flexibility of models when dealing with diverse graphs. 
For instance, node-wise positional encoding could force the representations of neighboring nodes to become similar, which may hurt the model performance in heterophily graphs where connected nodes tend to belong to different labels. 
Ideally, the constructed token sequences should encompass various graph information that reflects essential properties of the graphs.
Clearly, previous methods for constructing token sequences fall short in meeting this requirement.
Consequently, a natural question arises:
\textit{Could we design a new method to construct token sequences that comprehensively capture the complex graph features for each node?}

To answer this question, we present \tname, a new token sequence generator for graph Transformers.
Unlike previous methods that rely on single element-based token sequences as sentences to express the graph information of nodes,
\tname constructs a paragraph for each node's representation, 
which involves generating multiple types of sentences to characterize each target node.
By deeply analyzing the characteristics of different token elements (\ie node and neighborhood), \tname introduces two general token generators to extract complex graph features: the neighborhood-based token generator and node-based token generator.
Each token generator offers a flexible approach to construct token sequences from different feature spaces, ensuring preserve a wide range of properties comprehensively for nodes across diverse graphs.

Built upon \tname, we develop a composite \textbf{N}ode \textbf{T}okenized graph Trans\textbf{former} for node classification termed \name.
Based on the outputs of \tname, \name formulates a Transformer layer-based backbone to learn node representations from the generated token sequences.
Specifically, \name first leverages the Transformer layers to extract representations of different types of tokens, capturing various graph properties. 
Then, a feature fusion module is tailored to adaptively learn the final node representations from different token sequences to meet the unique requirement of each node on different graphs.
Since \tname can comprehensively capture necessary information from complex graph features, \name only  need to apply the standard Transformer layer for learning node representations, not requiring graph-specific modifications, such as tailored positional encoding or attention bias, ensuring the universality of \name.

Our main contributions are summarized as follows:
\begin{itemize}
    \item We present a novel token sequence generator, \tname, that generates token sequences based on various token elements from topology and attribute views. 

    \item We propose a new graph Transformer, \name, for node classification that utilizes standard Transformer layers and an adaptive fusion module to learn final node representations from the token sequences generated by \tname.

    \item We conduct extensive experiments on a diverse range of graph benchmark datasets, including both homophily and heterophily graphs. The results demonstrate the clear superiority of \name on the node classification task when compared to representative and powerful GNN-based methods.
\end{itemize}

\section{Related Work}


In this section, we first briefly review recent GNNs for the node classification task, then we introduce recent efforts for graph Transformers.

\subsection{Graph Neural Network}
In recent years, GNNs have achieved remarkable performance in  the task of node classification on graphs.
Classical GNNs develop various aggregation strategies to effectively extract meaningful information from local neighbors.
Graph convolutional network (GCN)~\cite{gcn}, as a pioneering work, introduces the first-order Laplacian smoothing to aggregate information of immediate neighbors.
Based on this typical model, many follow-up methods have been developed to enhance the model performance for node classification.
Some researches~\cite{jknet,mixhop,h2gnn} aim to introduce high-order neighborhood information into the aggregation process.   
Besides aggregating information on the original adjacency, certain researches~\cite{geomgcn,gdc,amgcn,spgcn} attempt to design aggregation operations on the constructed graphs.

Another line of GNNs introduce additional information~\cite{bmgcn} or techniques~\cite{gat} to guide the aggregation operation. 
Typical methods are the attention-based GNNs~\cite{gat,spgat,fagcn}, that tailor attention-based aggregation strategies, such as introducing self-supervised learning~\cite{spgat} and signed attention-based aggregation~\cite{fagcn}, to encode more valuable information into the model.
In addition to the traditional GNN architecture, recent methods~\cite{pt,rlp,appnp,sgc,gprgnn} decouple the operations of neighborhood aggregation and feature transformation in each GCN layer, resulting in two independent modules, named decoupled GCNs~\cite{pt}.
By removing redundant parameter matrices and activation functions, decoupled GCNs can effectively preserve information of multi-hop neighborhoods and exhibit competitive performance on the node classification task.

Despite the effectiveness of GNN-based methods, the inherent limitations of the message-passing mechanism, such as over-smoothing~\cite{oversmoothing} and over-squashing~\cite{oversq} issues, restrict GNNs from effectively preserving long-range dependencies in graphs, further hindering the model's potential in learning informative node representations.

\subsection{Graph Transformer}
We review recent graph Transformers from two categories according to the model input: entire graph-based graph Transformers and token sequence-based graph Transformers.

\textbf{Entire graph-based graph Transformers.}
Methods of this category require the whole graph as the model input.
Some methods~\cite{graphtrans,graphgps,sat,specformer} combine GNN layers and Transformer layers to construct a hybrid model, in which the GNN layer is used to preserve local topology features and the Transformer layer is used to preserve long-range information in graphs.
Besides combining GNNs, several studies augment the input node features with graph structural features (\eg Laplacian eigenvectors~\cite{gt,san}) or introduce the structure-aware attention bias~\cite{graphormer} into the Transformer architecture.
Another line of methods~\cite{nodeformer,ddiformer,treeatt,gapformer} focuses on reducing the computational overhead of the self-attention calculation by applying various linear self-attention calculation techniques~\cite{nodeformer} or pooling techniques~\cite{gapformer}.
Recently, Jinwoo \etal ~\cite{tokengt} propose TokenGT that treats nodes and edges of the input graph as independent tokens, resulting in a long sequence containing node-wise and edge-wise tokens.
Hence, we also regard TokenGT as the entire graph-based graph Transformer. 

\textbf{Token sequence-based graph Transformers.}
Methods of this category consider constructing a token sequence for each node as the model input, rather than the whole graph.
There are two types of strategies to construct the token sequence: node-based and neighborhood-based methods. 
Gophormer~\cite{gophormer} and ANS-GT~\cite{ansgt} utilize various node sampling strategies to sample the most relevant nodes to construct the token sequence for each target node. 
As a recent efficient work, NAGphormer~\cite{nagphormer} utilizes a novel module Hop2Token to transform multi-hop neighborhoods of the target into a token sequence.

Compared to entire graph-based graph Transformers, token sequence-based graph Transformers provide a more precise way to describe the complex graph features of nodes.
Nevertheless, token construction strategies in previous studies are limited to using a single token element (node or neighborhood) to generate the token sequences, leading to expressing partial graph information for each node.
A very recent study~\cite{vcrgt} also attempts to construct token sequences consisting of both node and neighborhood tokens.
However, it constructs one token sequence involving node and neighborhood tokens mainly based on the topology view, which inevitably limits the Transformer to flexibly learn node representations from complex graph features.

In contrast, our proposed \name first utilizes a new token generator \tname, which constructs the independent token sequences based on different token elements from topology and attribute views, comprehensively expressing the complex graph features for each node.
And \name further leverages a feature fusion module to adaptively extract node representations from different types of token sequences, ensuring the model's flexibility.

\section{Preliminaries}

\begin{figure*}[t]
\centering
\includegraphics[width=16cm]{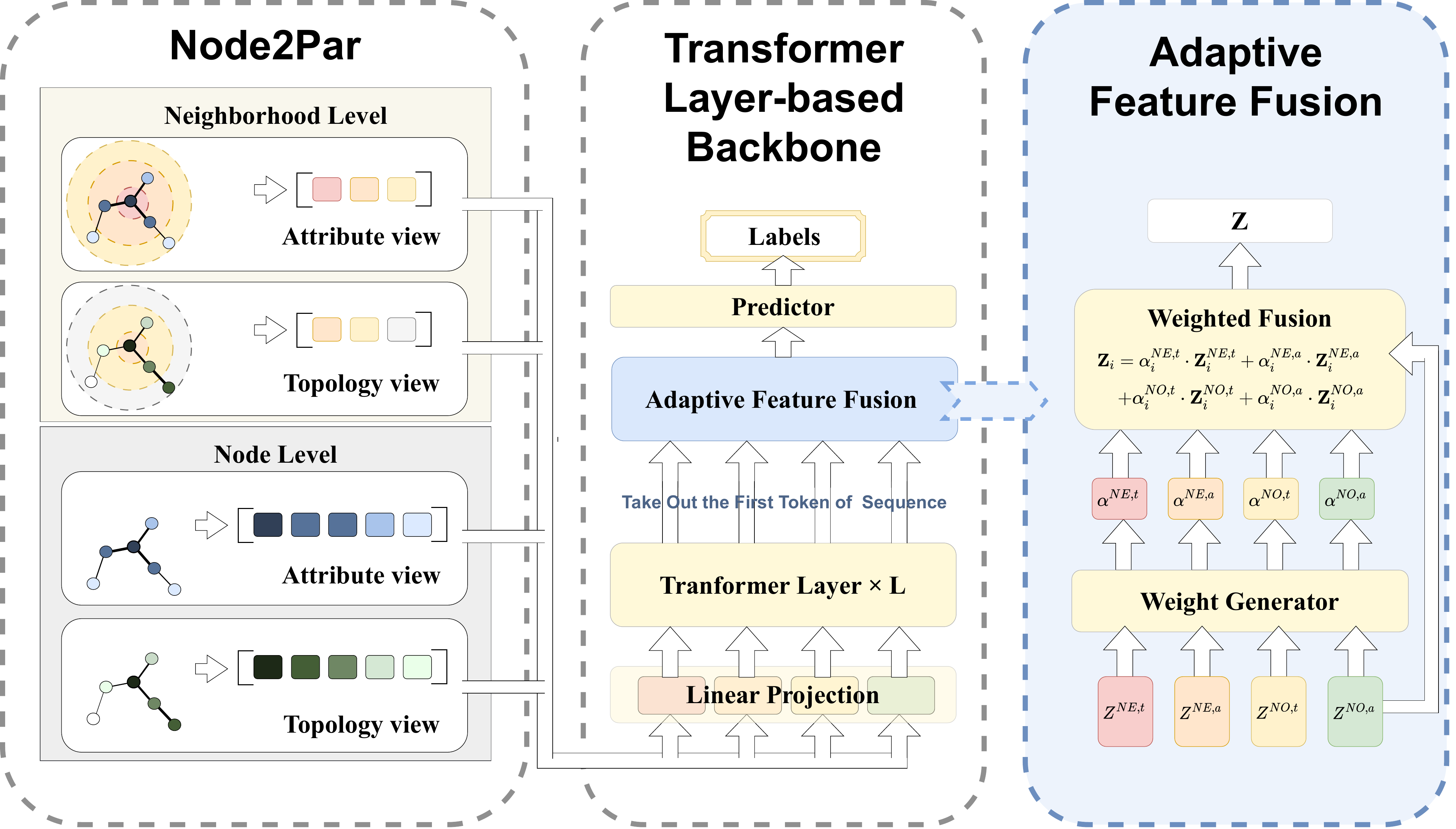}
\caption{
The overall framework of \name. Specifically, \name adopts the token sequences from different levels and views generated by \tname as the input.
Then, \name leverages a Transformer-based backbone with standard Transformer layers and the adaptive feature fusion module to learn final node representations from the constructed sequences.
}
\label{fig:method}
\end{figure*}

In this section, we first introduce the notations used in this paper, then we review typical architectures of GNNs and Transformers.

\subsection{Notations}
Given an attributed graph $\mathcal{G}=(V, E)$ where $V$ and $E$ are the sets of nodes and edges, we have the corresponding adjacency matrix $\mathbf{A}\in \mathbb{R}^{n\times n}$ ($n$ is the number of nodes).
If there is an edge between node $v_i$ and node $v_j$, $\mathbf{A}_{ij}=1$, otherwise $\mathbf{A}_{ij}=0$. 
The diagonal degree matrix $\mathbf{D}$ is defined as $\mathbf{D}_{ii}=\sum_{j=1}^n \mathbf{A}_{ij}$.
We also have the attribute feature matrix $\mathbf{X} \in \mathbb{R}^{n\times d}$ of nodes, where $d$ denotes the dimension of the attribute feature vector.
Each node is associated with a one-hot vector to represent the label information.
We use $\mathbf{Y}\in \mathbb{R}^{n\times c}$ to represent the label matrix of nodes, where $c$ is the number of labels.
Given a set of nodes $V_l$ whose labels are known, the goal of node classification is to predict the labels of the rest nodes in $V-V_l$.

\subsection{Graph Convolutional Network}
GCN~\cite{gcn} is a typical method of GNN that leverages the normalized adjacency matrix to aggregate information of immediate neighbors.
A GCN layer is defined as follows:
\begin{equation}
    \mathbf{H}^{(l+1)} = \sigma(\hat{\mathbf{A}}\mathbf{H}^{(l)}\mathbf{W}^{(l)}), 
    \label{eq:gcn}
\end{equation}
where $\hat{\mathbf{A}} = (\mathbf{D}+\mathbf{I})^{-\frac{1}{2}}(\mathbf{A}+\mathbf{I})(\mathbf{D}+\mathbf{I})^{-\frac{1}{2}}$ is the normalized adjacency matrix with self-loops, $\mathbf{I} \in \mathbb{R}^{n\times n}$ is the identity matrix.
$\mathbf{H}^{(l)}\in \mathbb{R}^{n\times d^{(l)}}$ and $\mathbf{W}^{(l)}\in \mathbb{R}^{d^{(l)}\times d^{(l+1)}}$
denote the node representations of nodes and the parameter matrix in the $l$-th GCN layer.
$\sigma(\cdot)$ denotes the nonlinear activation function.

\subsection{Transformer}
Here, we introduce the design of the Transformer layer, which is a key component of existing graph Transformers.
A Transformer layer consists of two core components: multi-head self-attention (MSA) and feed-forward networks (FFN).

Suppose the input feature matrix is $\mathbf{H} \in \mathbb{R}^{n\times d}$.
MSA first leverages projection layers to obtain the hidden representations, $\mathbf{Q}=\mathbf{H}\mathbf{W}^{Q}$, 
$\mathbf{K}=\mathbf{H}\mathbf{W}^{K}$,
$\mathbf{V}=\mathbf{H}\mathbf{W}^{V}$, where 
$\mathbf{W}^{Q}\in \mathbb{R}^{d\times d_k}$, 
$\mathbf{W}^{K}\in \mathbb{R}^{d\times d_k}$,
$\mathbf{W}^{V}\in \mathbb{R}^{d\times d_v}$ are projection matrices.
Then, the output of MSA is calculated as:
\begin{equation}
    \mathrm{MSA}(\mathbf{H}) = \mathit{softmax}(\frac{\mathbf{Q}\mathbf{K}^{\mathrm{T}}}{\sqrt{d_k}})\mathbf{V}.
    \label{eq:self-attention}
\end{equation}
The above equation illustrates the calculation of single-head attention, we omit its multi-head version for simplicity.  

FFN is composed of two fully connected layers and one nonlinear activation function, calculated as follows:
\begin{equation}
    \mathrm{FFN}(\mathbf{H}) = \mathrm{FC}(\sigma(\mathrm{FC(\mathbf{H})})),
    \label{eq:ffn}
\end{equation}
where $\mathrm{FC}(\cdot)$ denotes the fully connected layer.

\section{The Proposed Method}

In this section, we detail our proposed \name, which contains two components: \tname and Transformer layer-based backbone.
We first introduce \name, which involves two types of token generators that construct both neighborhood-based and node-based token sequences in different views for each target node.
Then, we introduce the Transformer layer-based backbone, which is tailored to learn node representations from different types of token sequences.
\autoref{fig:method} shows the overall framework of \name.

\subsection{Neighborhood-based Token Generator}
Neighborhood is an important element of graphs, which describes connections surrounding the target node in a certain range.
The key point of generating neighborhood-based token sequences is to choose a suitable approach to express the features of the neighborhood.
Previous studies utilize propagation-based strategies, such as random walk~\cite{nagphormer} or personalized PageRank~\cite{pamt,ncn} to calculate features of the neighborhood.
These strategies only effectively preserve the neighborhood information in the topology view, failing in capturing the neighborhood information in the attribute view, resulting in expressing partial neighborhood information.

To flexibly capture the neighborhood information in different feature views, we provide a general form of generating neighborhood-based tokens:
\begin{equation}
    \mathbf{X}^{N,(k)}_{i} = \sum_{v_j \in \mathcal{N}^{k}_{v_i}} \mathbf{W}^{(k)}_{ij} \cdot \mathbf{X}_{j},
    \label{eq:general_neighborhood}
\end{equation}
where $\mathcal{N}^{k}_{v_i}$ denotes the $k$-hop neighborhood of node $v_i$.
$\mathbf{W}^{(k)}_{ij}$ denotes the aggregation weight of node $v_j$ in $\mathcal{N}^{k}_{v_i}$ and $\mathbf{X}_{j}$ denotes the attribute feature of node $v_j$.
$\mathbf{X}^{N,(k)}_{i}$ denotes the feature of  $k$-hop neighborhood of node $v_i$.
$\mathbf{X}^{N,(0)}_{i} = \mathbf{X}_{i}$ means that we regard node $v_i$ itself as its 0-hop neighborhood.

Based on \autoref{eq:general_neighborhood}, we can observe that the feature of $k$-hop neighborhood is the weighted aggregation of node attribute features in the corresponding neighborhood.
Obviously, the aggregation weights determine how the aggregated features express the neighborhood information.
The aggregation weights in previous methods~\cite{appnp,gprgnn,nagphormer} are determined by the topology information (\ie the adjacency matrix), which could be regarded as expressing the neighborhood information in the topology view.

\begin{figure}[t]
\centering
\includegraphics[width=8.8cm]{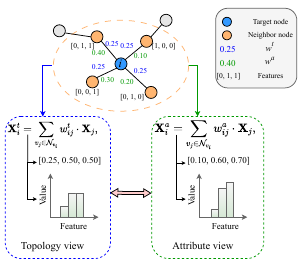}
\caption{
A toy example for illustrating the difference of neighborhood tokens generated from topology and attribute views.
Each node is associated with a three-dimensional feature vector.
We can clearly observe the difference between the two neighborhood tokens constructed from different views.
}
\label{fig:neighbor}
\end{figure}

Besides the topology view, attribute is also an important view for expressing the neighborhood information on attributed graphs. 
Intuitively, the neighborhood information from different views can provide different ways to describe the complex connections on the graph. 
\autoref{fig:neighbor} depicts the differences in neighborhood information obtained from different views.
Apparently, constructing the neighborhood-based token sequence from a single information view leads to incomplete expression of neighborhood information, further limiting the model performance.

Motivated by the above analysis, we construct two types of neighborhood-based token sequences for each node from both topology and attribute views.
Specifically, for the $k$-hop neighborhood of node $v_i$, the neighborhood features of different views are calculated as follows:
\begin{equation}
      \mathbf{X}^{t,(k)}_{i} = \sum_{v_j \in \mathcal{N}^{k}_{v_i}} \mathbf{W}^{t,(k)}_{ij} \cdot \mathbf{X}_{j}, 
    \label{eq:t_nei}   
\end{equation}
\begin{equation}
      \mathbf{X}^{a,(k)}_{i} = \sum_{v_j \in \mathcal{N}^{k}_{v_i}} \mathbf{W}^{a,(k)}_{ij} \cdot \mathbf{X}_{j},
    \label{eq:a_nei}   
\end{equation}
where $\mathbf{X}^{t,(k)}_{i}$ and $\mathbf{X}^{a,(k)}_{i}$ are the neighborhood tokens constructed under the views of topology and attribute, respectively.
$\mathbf{W}^{t,(k)}_{ij}$ and $\mathbf{W}^{a,(k)}_{ij}$ are the corresponding aggregation weights.
In practice, $\mathbf{W}^{t,(k)}_{ij}$ and $\mathbf{W}^{a,(k)}_{ij}$ are calculated as follows:
\begin{equation}
    \mathbf{W}^{t,(k)}_{ij} = \mathbf{A}^{t,(k)}_{ij}, ~~~~~~ \mathbf{A}^{t,(k)} = \hat{\mathbf{A}}^{k},  
    \label{eq:w_t}   
\end{equation}
\begin{equation}
     \mathbf{W}^{a,(k)}_{ij} = \mathbf{A}^{a,(k)}_{ij},~~~~~~ \mathbf{A}^{a,(k)} = (\mathbf{A} \odot \mathbf{A}^{s}) ^{k}, 
    \label{eq:w_a}   
\end{equation}
where $\mathbf{A}^{s}$ is the attribute similarity matrix calculated by $\mathrm{Cosine}(\mathbf{X},\mathbf{X}^{\mathrm{T}})$, measuring the cosine similarity of the node attribute features.

Finally, we generate the two types of neighborhood-based token sequences $S^{NE,t}_{i} = \{ \mathbf{X}^{t,(0)}_{i}, \dots, \mathbf{X}^{t,(K)}_{i}\}$
and $S^{NE,a}_{i} = \{ \mathbf{X}^{a,(0)}_{i}, \dots,  \mathbf{X}^{a,(K)}_{i}\}$ for node $v_i$ to express the neighborhood information from both topology and attribute views, where $K$ is a hyper-parameter representing the max range of neighborhood.

\subsection{Node-based Token Generator}
While neighborhood-based tokens can carefully preserve the structural information of graphs, they have limitations in capturing the node-level information, such as long-range dependencies due to the over-squashing issue~\cite{oversq}.
The lack of necessary node-level information can result in suboptimal node representations~\cite{vcrgt}.
To address this limitation, node-based tokens are introduced as a solution.

The key idea of constructing node-based token sequences is to sample relevant nodes which capture essential node-level information for each target node.
In this paper, we exploit a general strategy to sample nodes that contains two steps: 
(1) Measure the similarity scores of nodes; 
(2) Select the most similar nodes to construct the token sequence:
\begin{equation}
      S^{NO}_{i} = \{\mathbf{X}_j | v_j \in \mathrm{Top}(\mathbf{M}_{i})\},
    \label{eq:general_node}   
\end{equation}
where $\mathbf{M}\in \mathbb{R}^{n \times n}$ is the score matrix of all node pairs.
$\mathrm{Top}(\cdot)$ represents that we select $n_k$ nodes with top similarity scores, $n_k$ is a hyper-parameter.
$S^{NO}_{i}$ is the node-based token sequence of node $v_i$.

Similar to neighborhood-based tokens, there are also two views to measure the similarity of each node pair: topology view and attribute view.
Different views measure the similarity of nodes in different feature spaces, which can result in different node candidates.
Previous studies~\cite{amgcn,simpgcn} have revealed that nodes with highly similar attribute features may not possess close relations in topology structure. 
This situation implies that constructing node-based token sequences based on the node similarity in a single feature space is inefficient in capturing complex relations of nodes in the graph.
Hence, in this paper, we generate two types of node-based token sequences based on the topology features and attribute features of nodes, respectively.

To exploit the topology features to estimate the similarity matrix $\mathbf{M}^{t}$, we adopt the personalized PageRank (PPR) method to calculate the similarity scores of nodes.
Specifically, for the target node $v_i$, the PageRank scores is calculated as follows:
\begin{equation}
    \mathbf{s}_{i}^{(l)} = r\cdot \hat{\mathbf{A}}\mathbf{s}_{i}^{(l-1)} + (1-r) \cdot \mathbf{s}^{i}_{0},
    \label{eq:pprscore}
\end{equation}
where $\mathbf{s}_{i}^{(l)} \in \mathbb{R}^{n\times 1}$ denotes the PPR scores of all nodes from the target node $v_i$ at the $l$-th propagation step and $\mathbf{s}_{i}^{(0)} = \hat{\mathbf{A}}_{i}$.
$r$ is the damping constant factor and $\mathbf{s}^{i}_{0} \in \mathbb{R}^{n\times 1}$ is the one-hot personalized vector where the element corresponding to the target node equals 1.
In practice, we utilize a two-step propagation to estimate the PPR scores of nodes.
Finally, PPR scores of all nodes constitute the similarity matrix $\mathbf{M}^{t}$.
As for estimating the attribute-aware similarity matrix $\mathbf{M}^{a}$, we directly apply the cosine similarity to calculate the similarity scores of nodes based the raw attribute features according to the empirical results in previous methods~\cite{amgcn,simpgcn}.

According to different score matrices $\mathbf{M}^{t}$ and $\mathbf{M}^{a}$, for node $v_i$, we can obtain the corresponding token sequences $S^{NO,t}_{i}$ and $S^{NO,a}_{i}$ via \autoref{eq:general_node}.
Similar to neighborhood-based token sequences, we also add the node itself as the first token of $S^{NO,t}_{i}$ and $S^{NO,a}_{i}$ to enable the target node to learn valuable graph information from the constructed token sequences.

\subsection{Transformer-based Backbone}
Through the proposed token generators, we obtain four token sequences $\{{S^{NE,t}_{i}}, {S^{NE,a}_{i}}, {S^{NO,t}_{i}}, {S^{NO,a}_{i}}\}$ for node $v_i$.
We further formulate a Transformer layer-based backbone, which contains a series of standard Transformer layers and an adaptive feature fusion layer, to learn the representation of node $v_i$ from the constructed token sequences.

First, we leverage the standard Transformer layer to learn the representations of node $v_i$ from the corresponding token sequence.
Take the token sequence ${S^{NE,t}_{i}}$ for instance, we first utilize a project layer to obtain the input features:
\begin{equation}
      \mathbf{H}^{NE,t^{(0)}}_{i} = [\mathbf{X}^{t,(0)}_{i}\mathbf{W}^{p}, \dots, \mathbf{X}^{t,(K)}_{i}\mathbf{W}^{p}],
    \label{eq:input}   
\end{equation}
where $\mathbf{W}^{p} \in \mathbb{R}^{d \times d^{(0)}}$ is the projection matrix and $\mathbf{H}^{NE,t^{(0)}}_{i} \in \mathbb{R}^{(K+1) \times d^{(0)}}$ is the input feature matrix regarding the token sequence ${S^{NE,t}_{i}}$.
A series of standard Transformer layers are adopted to learn representations of node $v_i$ from the input $\mathbf{H}^{NE,t^{(0)}}_{i}$:
\begin{equation}
      \mathbf{H}^{NE,t^{(l)\prime}}_{i} = \mathrm{MSA}(\mathbf{H}^{NE,t^{(l)}}_{i}) + \mathbf{H}^{NE,t^{(l)}}_{i},
    \label{eq:sa_token}   
\end{equation}
\begin{equation}
      \mathbf{H}^{NE,t^{(l+1)}}_{i} = \mathrm{FFN}(\mathbf{H}^{NE,t^{(l)\prime}}_{i}) + \mathbf{H}^{NE,t^{(l)\prime}}_{i},
    \label{eq:ffn_token}   
\end{equation}
where $\mathbf{H}^{NE,t^{(l+1)}}_{i}$ denotes the output of the $(l+1)$-th Transformer layer.
Suppose $\mathbf{H}^{NE,t}_{i}\in \mathbb{R}^{(K+1)\times d^{o}}$ is the final output of the Transformer layer, we take out the first row of $\mathbf{H}^{NE,t}_{i}$ denoted as $\mathbf{Z}^{NE,t}_{i} \in \mathbb{R}^{1 \times d^{o}}$ to represent the node representation of $v_i$ learned from the token sequence ${S^{NE,t}_{i}}$. 
From the same processing, we can obtain representations learned from other token sequences, $\mathbf{Z}^{NE,a}_{i}$, $\mathbf{Z}^{NO,t}_{i}$ and $\mathbf{Z}^{NO,a}_{i}$.

The above representations are the hidden features of complex graph information from different token elements under different views.
As for the final node representation, considering that the properties of different graphs vary largely, an ideal solution is to adaptively fuse the above features according to various demands of nodes on various graphs~\cite{ncn}. 
To this end, we develop a feature fusion layer to obtain the final node representation from representations learned from different types of token sequences.

Specifically, we first calculate the fusion weights for each learned representation:
\begin{equation}
    \alpha^{NE,t}_{i} = \sigma (\mathbf{Z}^{NE,t}_{i} \cdot \mathbf{W}^{f_0} ) \cdot \mathbf{W}^{f_1},  
    \label{eq:fusion_token}   
\end{equation}
where $\mathbf{W}^{f_0}\in \mathbb{R}^{d^o \times d^f}$ and $\mathbf{W}^{f_1} \in \mathbb{R}^{d^f \times 1}$ are
two learnable parameter matrices.
$\sigma(\cdot)$ is the activation function and $\alpha^{NE,t}_{i}$ is a scalar value representing the aggregation weight.

Similarly, we can obtain $\alpha^{NE,a}_{i}$, $\alpha^{NO,t}_{i}$, $\alpha^{NO,a}_{i}$ of the corresponding representations. 
We further leverage the softmax function to normalize these weights.
The final representation of $v_i$ is calculated as follows:
\begin{equation}
\begin{split}
     \mathbf{Z}_{i}  & = \alpha^{NE,t}_{i} \cdot \mathbf{Z}^{NE,t}_{i} + \alpha^{NE,a}_{i} \cdot \mathbf{Z}^{NE,a}_{i} \\
     & + \alpha^{NO,t}_{i} \cdot \mathbf{Z}^{NO,t}_{i} + \alpha^{NO,a}_{i} \cdot \mathbf{Z}^{NO,a}_{i},
    \label{eq:final_representation}    
\end{split}
\end{equation}
where $\mathbf{Z}_{i}$ is the final representation of $v_i$.
In this way, \name can adaptively learn node representations from various token sequences.

As for the node classification task, we leverage the MLP with two linear layers as the predictor to predict node labels $\hat{\mathbf{Y}}$ and adopt the Cross-entropy loss function to train the model parameters:
\begin{equation}
\mathcal{L} = -\sum_{i\in V_{l}} {\mathbf{Y}_{i}}\mathrm{ln}\hat{\mathbf{Y}}_{i}, ~~~~\hat{\mathbf{Y}} = \mathrm{MLP}(\mathbf{Z}).
    \label{eq:loss}   
\end{equation}

\subsection{Discussions on \name}
Here, we provide discussions on \name about the relations with the previous token sequence-based graph Transformers.

Existing graph Transformers~\cite{gophormer,ansgt,nagphormer} only construct node-oriented~\cite{gophormer,ansgt} or neighborhood-oriented~\cite{nagphormer} token sequences,
which could be regarded as a part of \tname.
NAGphormer and Gophormer construct the neighborhood-wise and node-wise token sequences from the topology view, respectively.
While ANS-GT mixes the node-wise token sequences from topology and attribute views through a reinforcement learning-based strategy. 
Since the constructed token sequences of previous methods just reflect partial graph information, these methods need to introduce specific graph information into the Transformer architecture, such as structural attention biases~\cite{gophormer,ansgt}, resulting in confined applicability. 

In contrast, our proposed \name generates token sequences from both neighborhood- and node-oriented from different views through \tname, which could preserve more valuable graph information than previous methods.
Consequently, \name requires no additional specific graph information and just leverages the standard Transformer layer to learn node representations, guaranteeing the generalization on different types of graphs.

Moreover, \name is a flexible method.
Any reasonable strategy of generating $\mathbf{W}^{(k)}$ and $\mathbf{M}$ could be adopted to construct new types of neighborhood-based and node-based token sequences in \tname according to specific application scenarios, respectively.

\section{Experiments}
In this section, we first introduce the experimental setup, containing datasets and baselines.
Then, we exhibit the results of conducted experiments, including performance comparison, ablation studies, and parameter sensitivity analysis, which are helpful to understand our proposed method deeply.

\subsection{Datasets}
We adopt 12 widely used benchmark graphs extracted from various domains, including research paper-based graphs, purchase action-based graphs, social relation-based graphs, and web page-based graphs.

\textbf{Research paper-based graphs.} 
Pubmed~\cite{nagphormer}, Citeseer~\cite{amgcn}, Corafull~\cite{nagphormer} and UAI2010~\cite{amgcn} fall in this category.
Nodes represent the research paper and edges represent the citation or co-author relations between papers in these graphs. 
The attribute features of nodes are generated by the title, abstract, and introduction of the research paper.
The labels of nodes represent research areas of papers.

\textbf{Purchase action-based graphs.} 
Photo~\cite{nagphormer}, Computer~\cite{nagphormer} and Amazon2M~\cite{nagphormer} are co-purchase networks extracted from the Amazon website.
Nodes represent goods. On Photo and Computer, edges indicate that two goods appear together in the same shopping list and on Amazon2M edges signify that the connected goods are frequently bought together.
Features of nodes are generated by the reviews of goods.
Labels represent the categories of goods.

\textbf{Social relation-based graphs.} 
BlogCatalog \cite{amgcn}, Flickr \cite{amgcn} and Reddit \cite{nagphormer} are three social networks extracted from the famous social platforms, BlogCatalog and Flickr~\cite{heterdataset}.
Nodes represent users, and edges represent the social relations between users in these graphs.
Node features are generated by the information of user profiles.
Labels denote the interest category of a user.

\textbf{Web page-based graphs.}
Squirrel~\cite{heterdataset} is a web page-based graph where nodes represent Wikipedia pages and edges represent links between pages.
The node features are generated by the content of the corresponding web page.
The node label denotes the average monthly traffic.
Note that the original graph contains too much duplicate nodes~\cite{heterdataset}.
Hence, we adopt the filtered version proposed by~\cite{heterdataset} for experiments.

\textbf{Word dependency graphs.}
Roman-empire\cite{heterdataset} is based on the Roman Empire article from English Wikipedia.
Each node in the graph corresponds to one (non-unique) word in the text.
Two words are connected with an edge if at least one of the following two conditions holds: either these words follow each other in the text, or these words are connected in the dependency tree of the sentence (one word depends on the other).
The node features are generated using FastText word embeddings, and each node is classified by its syntactic role.

\begin{table}[]
    \centering
	\caption{Statistics on datasets. Rank by the number of nodes from small to large.}
	\label{tab:dataset}
 \renewcommand\arraystretch{1.1}
\scalebox{0.91}{
	\begin{tabular}{lrrrrcc}
		\toprule
		Dataset & \# Nodes $\uparrow$ & \# Edges & \# Features & \# Classes & $\mathcal{H}$ \\
		\midrule
        Squirrel& 2,223& 46,998& 2,089 & 5 & 0.21  \\
        UAI2010 & 3,067 & 28,311 & 4,973 & 19 & 0.36\\
        Citeseer & 3,327 & 4,732 & 3,703 & 6  & 0.74\\
        BlogCatalog & 5,196 & 171,743 & 8,189 & 6& 0.40\\
        Flickr & 7,575 & 239,738 & 12,047 & 9& 0.24\\
        Photo & 7,650 & 238,163 & 745 & 8 & 0.83\\

        Computer & 13,752 & 491,722 & 767 & 10& 0.78\\

        Corafull & 19,793 & 126,842 & 8,710 & 70& 0.57\\
        Pubmed & 19,717 & 44,324 & 500 & 3  & 0.80\\

        Roman-empire&22,662&32,927&300&18&0.05\\
        Reddit&232,965&11,606,919&602&41&0.78\\

        Amazon2M &2,449,029&51,859,140&100&47&0.81\\

		\bottomrule
	\end{tabular}}
\end{table}


The edge homophily ratio~\cite{glognn} $\mathcal{H}(\mathcal{G}) \in [0, 1]$ is adopted to measure the homophily degree of a graph.
${H}(\mathcal{G}) \rightarrow 1$ means strong homophily, which implies that connected nodes tend to belong to the same label.
While low ${H}(\mathcal{G}) \rightarrow 0$ means strong heterophily, which means that connected nodes tend to belong to different labels.
The detailed statistics of datasets are reported in \autoref{tab:dataset}.

\subsection{Baselines}
We select 11 representative methods for the node classification task, including GNNs and graph Transformers as the baselines. 
The detailed descriptions of adopted baselines are as follows.
\begin{itemize}
    \item GCN~\cite{gcn} is a classical graph neural network that leverages the first-order Laplacian smoothing operation to aggregate the information of immediate neighbors.

    \item SGC~\cite{sgc} is a simplifying version of GCN that removes the nonlinear activation function between GCN layers.

    \item APPNP~\cite{appnp} first leverages the MLP module to obtain hidden representations of nodes and then utilizes the personalized PageRank to aggregate the information of multi-hop neighborhoods.

    \item GPRGNN~\cite{gprgnn} is an improved version of APPNP, which leverages the learnable aggregation weights to fuse the information of multi-hop neighborhoods, resulting in more expressive node representations. 

    \item FAGCN~\cite{fagcn} proposes a signed attention-based aggregation mechanism to aggregate different-frequency information of immediate neighbors.

    \item NodeFormer~\cite{nodeformer} is a recent graph Transformer for the node classification task that leverages the linear self-attention mechanism to learn the node representations. Moreover, learnable structural biases are introduced to enhance the model performance.

    \item SGFormer~\cite{sgformer} is a hybrid model that combines a novel linear self-attention layer and GNN-style modules to learn node representations from the input graph.

    \item ANS-GT~\cite{ansgt} is a representative node token-based graph Transformer. It first sifts nodes based on topology and attribute features. Then, ANS-GT introduces reinforcement learning to guide the token sequence generation. In addition, many graph-specific strategies have been developed, such as global virtual nodes and structural-aware attention bias, to improve the model performance.

    \item NAGphormer~\cite{nagphormer} is a representative neighborhood token-based graph Transformer. It develops a  Hop2Token module to generate the neighborhood token via the propagation method. While the node-wise positional encoding is adopted as the model input. 

    \item Specformer~\cite{specformer} utilizes the Transformer layer to capture the interactions of different eigenvalues of the graph Laplacian matrix to fully preserve the different-frequency information of the input graph.

    \item STAGNN~\cite{sta} is an updated version of GPRGNN, which introduces the self-attention mechanism to measure the different contributions of nodes to the corresponding neighborhood.

\end{itemize}

\subsection{Parameter settings}
We perform hyper-parameter tuning for each baseline based on the recommended settings in the official implementations.
For the proposed \name, in the data preprocessing process,
we try the maximum range of neighborhood in $\{2, 3, 10, 15$\} and the number of selected nodes in $\{5, 10, 15$\}. 
For the model configuration of \name, we try the hidden dimension in $\{512, 768, 1024, 2048$\}, the intermediate dimension of the feature fusion layer in $\{64, 128, \dots, 1024$\}, the number of Transformer layers in $\{1, 2, 3\}$. 
Parameters are optimized with AdamW~\cite{adamw}, using a learning rate of in $\{1e-4, 2e-4, 2.5e-4, 5e-4\}$. 
We also search the dropout rate in $\{0.1, 0.3\}$, the attention dropout in $\{0.5, 0.7, 0.9\}$ and attention heads in $\{1, 8\}$.
Besides, the batch size is set to 1024, the weight decay is set to 1e-05, and the training process is early stopped within 50 epochs.
All experiments are conducted on a Linux server with one Xeon Silver 4210 CPU and four RTX TITAN GPUs. 
Codes are implemented on Python 3.8 and Pytorch 1.11. 

\begin{table*}[t]

  \centering
      \caption{
Comparison of all models in terms of mean accuracy $\pm$ stdev (\%) on small-scale datasets. The best results appear in \textbf{bold}. The second results appear in \underline{underline}.
    	}
  \label{tab:nc_results}
   \renewcommand\arraystretch{1.2}
  \scalebox{0.94}{
  \begin{tabular}{lcccccccccc}
    \toprule 
    \textbf{Method} & \textbf{Photo} 
    & \textbf{Pubmed}& \textbf{Computer}
    &\textbf{Citeseer}&\textbf{Corafull}&\textbf{BlogCatalog}
    &\textbf{UAI2010}&\textbf{Flickr}&\textbf{Squirrel}
    &\textbf{Roman-empire}\\
    $\mathcal{H}(G)$& 0.83 & 0.80 & 0.78 & 0.74 & 0.57 & 0.40& 0.36 & 0.24 & 0.21 & 0.05\\ 
    \midrule 
    GCN & 94.67\tiny{$\pm$0.22} & 86.69\tiny{$\pm$0.17} & 89.59\tiny{$\pm$0.07} & 80.33\tiny{$\pm$0.88} & 68.08\tiny{$\pm$0.12} & 79.04\tiny{$\pm$0.26} & 71.09\tiny{$\pm$0.50}& 80.97\tiny{$\pm$0.51}& 36.88\tiny{$\pm$1.80} & 60.82\tiny{$\pm$0.14}\\

    APPNP & 95.73\tiny{$\pm$0.12} & 88.60\tiny{$\pm$0.06} & 90.93\tiny{$\pm$0.18} & 80.93\tiny{$\pm$0.48} & 67.14\tiny{$\pm$0.14} & 95.18\tiny{$\pm$0.31} & 77.77\tiny{$\pm$0.68}& 92.28\tiny{$\pm$0.23}& 36.69\tiny{$\pm$1.82} & 51.84\tiny{$\pm$0.18}\\

    SGC & 93.40\tiny{$\pm$0.23} & 79.51\tiny{$\pm$0.04} & 89.09\tiny{$\pm$0.10} & 74.22\tiny{$\pm$0.19} & 54.52\tiny{$\pm$0.44} & 73.69\tiny{$\pm$0.13} & 71.32\tiny{$\pm$0.84}& 80.75\tiny{$\pm$0.17}& 33.90\tiny{$\pm$1.10} & 40.33\tiny{$\pm$0.34} \\


    GPRGNN & 95.22\tiny{$\pm$0.35} & 87.82\tiny{$\pm$0.25} & 90.50\tiny{$\pm$0.91} & 81.02\tiny{$\pm$0.71} & 72.55\tiny{$\pm$0.75} & \underline{95.66\tiny{$\pm$0.17}} & 77.23\tiny{$\pm$0.49}& \underline{92.53\tiny{$\pm$0.67}}& 39.01\tiny{$\pm$0.96}& 67.04\tiny{$\pm$0.92} \\

    FAGCN & 94.51\tiny{$\pm$0.44} & 88.28\tiny{$\pm$0.16} & 82.77\tiny{$\pm$1.19} & 79.28\tiny{$\pm$1.70} & 69.39\tiny{$\pm$0.93} & 94.41\tiny{$\pm$0.32} & 74.71\tiny{$\pm$0.54}& 92.13\tiny{$\pm$0.31}& \underline{39.43\tiny{$\pm$0.37}} & 75.71\tiny{$\pm$1.39} \\

    \hline
    
    NodeFormer & 94.30\tiny{$\pm$0.55} & 88.27\tiny{$\pm$0.04} & 88.11\tiny{$\pm$0.29} & 78.56\tiny{$\pm$0.09} & 63.06\tiny{$\pm$0.86} & 93.33\tiny{$\pm$0.85} & 73.87\tiny{$\pm$1.39}& 90.39\tiny{$\pm$0.49}& 34.35\tiny{$\pm$1.31} & 58.37\tiny{$\pm$1.19}\\
    
    ANS-GT & 94.98\tiny{$\pm$0.43} & 89.14\tiny{$\pm$0.26} & 90.17\tiny{$\pm$0.44} & 77.60\tiny{$\pm$0.51} & 66.42\tiny{$\pm$0.23} & 94.18\tiny{$\pm$0.28} & 76.40\tiny{$\pm$0.48}& 89.95\tiny{$\pm$0.79}& 34.46\tiny{$\pm$0.83} & 69.64\tiny{$\pm$1.16} \\

    SGFormer & 93.13\tiny{$\pm$0.87} & 89.22\tiny{$\pm$0.24} & 84.40\tiny{$\pm$0.82} & 77.98\tiny{$\pm$0.91} & 63.46\tiny{$\pm$1.31} & 94.32\tiny{$\pm$0.21} & 75.17\tiny{$\pm$0.49}& 89.11\tiny{$\pm$1.04}& 36.86\tiny{$\pm$1.74} & 65.39\tiny{$\pm$0.78} \\
    
    NAGphormer & \underline{95.94\tiny{$\pm$0.45}} & 89.39\tiny{$\pm$0.13} & \underline{91.18\tiny{$\pm$0.36}} & 80.57\tiny{$\pm$0.72} & 72.15\tiny{$\pm$0.59} & 95.47\tiny{$\pm$0.87} & 78.37\tiny{$\pm$0.83}& 91.21\tiny{$\pm$0.94}& 38.14\tiny{$\pm$1.79} & 74.32\tiny{$\pm$0.65}\\ 

    Specformer & 95.80\tiny{$\pm$0.26} & 89.47\tiny{$\pm$0.48} & 91.09\tiny{$\pm$0.27} & 79.38\tiny{$\pm$0.96} & \underline{72.60\tiny{$\pm$0.24}} & 95.61\tiny{$\pm$0.34} & 77.80\tiny{$\pm$0.70}& 92.33\tiny{$\pm$0.46}& 37.47\tiny{$\pm$0.12} & 62.73\tiny{$\pm$1.96} \\

    STAGNN & 94.47\tiny{$\pm$0.73} & \textbf{90.18\tiny{$\pm$0.26}} & 88.37\tiny{$\pm$1.78} & \underline{81.05\tiny{$\pm$0.82}} & 72.22\tiny{$\pm$0.35} & 95.53\tiny{$\pm$0.20} & \underline{79.19\tiny{$\pm$0.69}}& 92.46\tiny{$\pm$0.26}& 38.21\tiny{$\pm$2.07} & 
    \underline{76.71\tiny{$\pm$0.33}}\\

    \hline
    \name & \textbf{96.17\tiny{$\pm$0.29} }& \underline{89.78\tiny{$\pm$0.21}} & \textbf{91.69\tiny{$\pm$0.32}} & \textbf{81.61\tiny{$\pm$0.75}} & \textbf{73.46\tiny{$\pm$0.29}} & \textbf{96.22\tiny{$\pm$0.38}} & \textbf{80.03\tiny{$\pm$0.97}} & \textbf{92.90\tiny{$\pm$0.20}} & \textbf{40.72\tiny{$\pm$1.16}}& \textbf{78.39\tiny{$\pm$1.11}} \\ 
    
    \bottomrule 
  \end{tabular}}
\end{table*}

\subsection{Comparison on Small-scale Datasets}
We adopt the accuracy metric to evaluate the performance of each model on node classification.
Specifically, for each dataset, we run each model ten times with different random seeds and report the average accuracy and the corresponding standard deviation.
The results are summarized in \autoref{tab:nc_results}.

We can observe that \name achieves the best performance on all datasets except Pubmed (\name still exhibits very competitive performance), demonstrating the superiority of \name on the node classification task.
In addition, \name beats ANS-GT and NAGphormer on all datasets.
This phenomenon indicates that enhancing the quality of token sequences can significantly improve the model performance. 
Moreover, this situation also reveals that by generating expressive token sequences, the pure Transformer architecture without graph-specific modifications still possesses great potential for node classification. 

Another observation is that advanced graph Transformers that introduce new techniques to integrate GNN and Transformer, such as Specformer and STAGNN, achieve competitive performance compared to the hybrid framework-based graph Transformers that directly combine GNN and Transformer.
This phenomenon reveals the potential of graph Transformers on graph representation learning.
Moreover, this situation also implies that how to effectively incorporate GNN with Transformer is a promising way to enhance the performance of graph Transformers on graph mining tasks.

\begin{table}[]
    \centering
	\caption{
	Comparison of \name and four scalable graph Transformers in terms of mean accuracy $\pm$ stdev (\%) on large-scale datasets. 
	The best results appear in \textbf{bold}.
	}
	\label{tab:large-data}
\scalebox{1}{
	\begin{tabular}{lcc}
		\toprule
		Method  &Reddit &Amazon2M\\
		\midrule
		NodeFormer &   88.97 $\pm$ 0.32   & 71.56 $\pm$ 0.42  \\
        ANS-GT &   93.27 $\pm$ 0.22   & 76.32 $\pm$ 0.38  \\
		SGFormer & 89.63 $\pm$ 0.26  & 74.22 $\pm$ 0.36   \\
        NAGphormer   & 93.58 $\pm$ 0.05 & 77.43 $\pm$ 0.24\\
		\midrule
		\name & \textbf{93.75 $\pm$ 0.26} & \textbf{78.03 $\pm$ 0.44}\\
		\bottomrule
	\end{tabular}
        }
 	
\end{table}
\subsection{Comparison on Large-scale Datasets}
Handling the graph mining tasks on large-scale graphs is a crucial challenging for advanced graph Transformers.
To evaluate the scalability of \name on large-scale graphs, we extended our comparison to two popular large-scale datasets, Reddit and Amazon2M.
Here, we compare the performance of \name and four scalable graph Transformers since other methods meet the out-of-memory issue on these datasets.  
The performance are reported in \autoref{tab:large-data}.
We can observe that \name consistently outperforms recent graph Transformers on two datasets, indicating that \name is capable of effectively handling the node classification task on large-scale graphs.

\begin{figure*}[t]
\centering
\includegraphics[width=17cm]{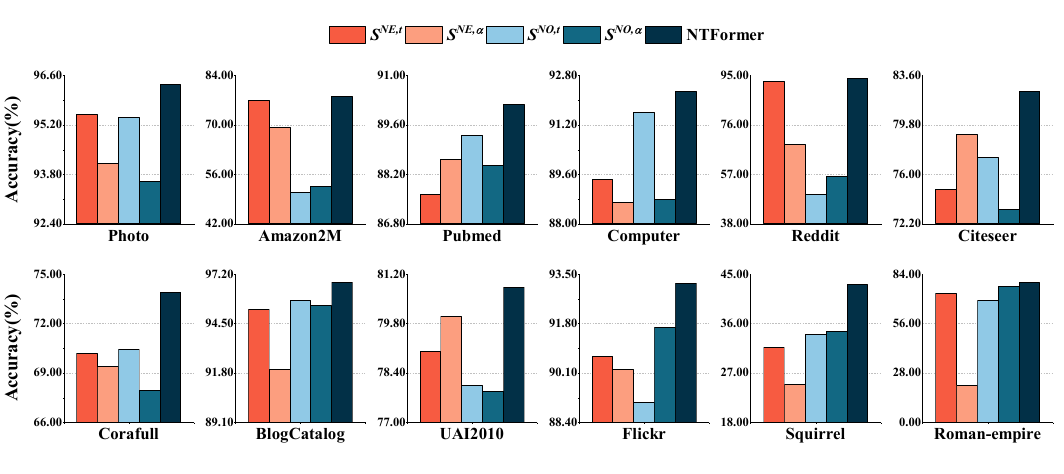}
\caption{
Performances of \name and its variants on all datasets.
}
\label{fig:ablation}
\end{figure*}

\begin{figure}[h]
\centering
\includegraphics[width=7cm]{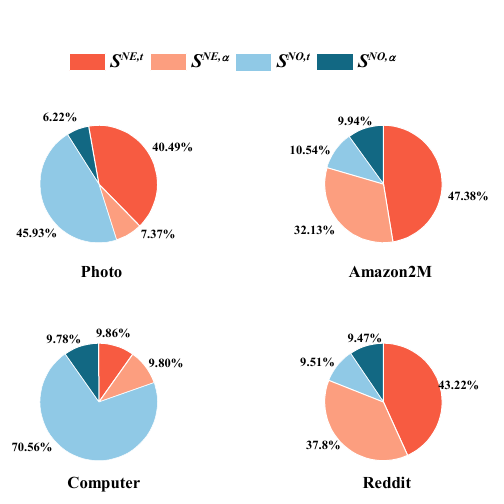}
\caption{
Visualization of the average weight proportion for each sequence on different graphs.
}
\label{fig:weight}
\end{figure}

\subsection{Study on Token Sequences}
\tname is the key module of \name that constructs four token sequences $\{{S^{NE,t}}, {S^{NE,a}}, {S^{NO,t}}, {S^{NO,a}}\}$ to describe various graph information for each node, which exhibits the significant difference with previous tokenized graph Transformers.
To evaluate the effectiveness of \tname, we design four variants of \name.
Each variant only leverages one token sequence from $\{{S^{NE,t}}, {S^{NE,a}}, {S^{NO,t}}, {S^{NO,a}}\}$ to learn node representations.
We further evaluate the performance of these variants on all datasets.
Results are shown in Figure \ref{fig:ablation}, from which 
we have the following observations:

(1) \name beats all four variants on all datasets, which indicates that combining token sequences constructed from different views can effectively enhance the model performance for node classification.

(2) Performances of four variants vary within the same dataset, implying that the contributions of these token sequences are different on the model performance. Naturally, we believe that the higher the accuracy of a sequence, the more important the information it provides.
We further visualize the adaptive fusion weights of different token sequences to validate whether \name can adaptively capture the varying importance of token sequences on different graphs.
We select four datasets involving small- and large-scale graphs for experiments.
Results are shown in Figure \ref{fig:weight}.
We observe a high consistency between the distribution of adaptive weights and the accuracy of node classification with a single sequence. This suggests that \name can adaptively learn graph-specific information by predicting the contributions of different token sequences through learning adaptive weights, further confirming the superiority of \name for learning node representations in different graphs.

(3) Influence of the same token sequence varies across different graphs, indicating that graphs with different characteristics prefer different token sequences.
Generally, homophily graphs prefer neighborhood-based token while heterophily graphs perform terrible without node-based token sequences.
This is because preserving local topological information is crucial for homophily graphs.
However, in heterophily graphs, due to the presence of numerous non-similar nodes within the neighborhood, the significance of neighborhood information diminishes.
Consequently, node-level information becomes especially critical.
This phenomenon also reveals that the graph information is very complex and hard to express by a single-type-based token sequence.
Hence, constructing token sequences from different elements and views for each node is necessary and crucial for improving the model performance on graph representation learning.

\begin{figure*}[t]
\centering
\includegraphics[width=17cm]{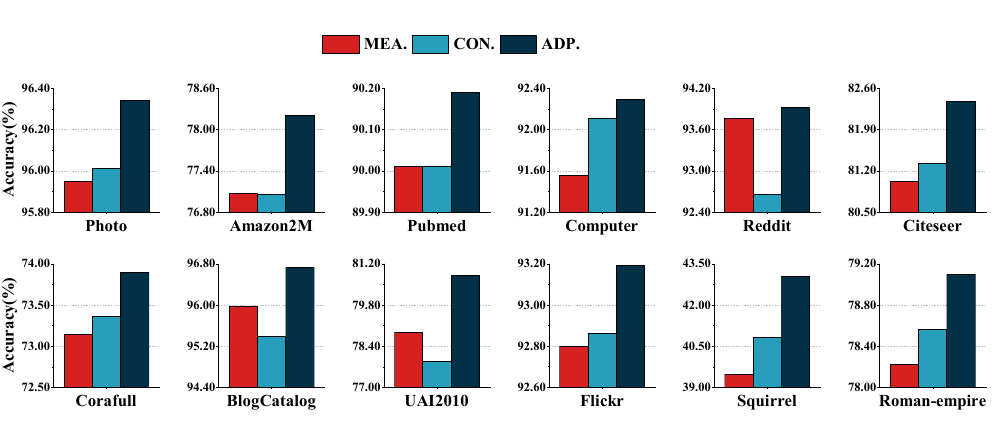}
\caption{
Performances of \name with different feature fusion strategies. 
MEA. denotes the average fusion.
CON. denotes the direct concatenation.
ADP. denotes the adaptive fusion.
}

\label{fig:readout}
\end{figure*}

\subsection{Study on Adaptive Fusion}
To obtain the final node representations for predicting the labels, \name develops an adaptive fusion module as a readout function to adaptively fuse the information from different token sequences.
By visualizing the weights in Figure \ref{fig:weight}, we have already validated the interpretability of the proposed readout function.
Here, we further validate the effectiveness of the adaptive fusion by comparing its performance with other common readout functions.
Specifically, we adopt two traditional readout functions: average fusion and concatenation fusion for experiments.
Average fusion indicates that \name uses equal weights to fuse the information of four token sequences.
Concatenation fusion denotes that \name directly combines the representations learned from different sequences via the vector concatenation operation.
We evaluate the performance of \name with different readout functions and the results are shown in \autoref{fig:readout}.
We can clearly observe that the adaptive fusion beats other strategies over all datasets, indicating that adaptively fusing information from different token sequences can significantly improve the model performance.
Results shown in \autoref{fig:weight} also explain this phenomenon since the contributions of different token sequences vary on different datasets.

\subsection{Parameter Sensitivity Analysis}
There are two key hyper-parameters in \name:
the maximum range of neighborhood $K$ and the number of sampling nodes $n_k$.
These parameters control the generation of token sequences, which further affect learning node representations.
Here, we conduct experiments to analyze the influence of these parameters on the model performance.
\begin{figure*}[t]
\centering
\includegraphics[width=17cm]{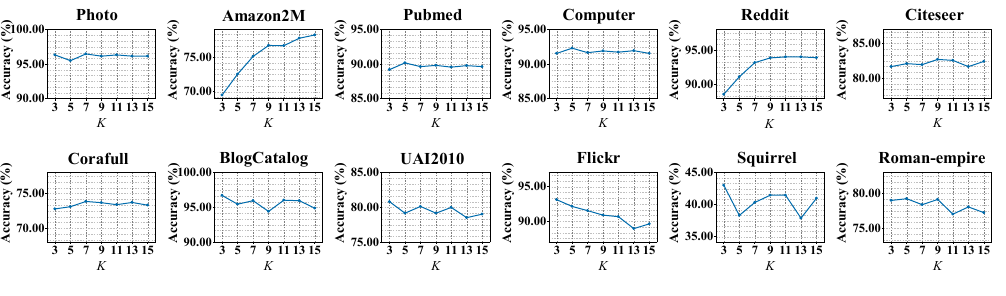}
\caption{
Performance of \name with different $K$ on all datasets.
}
\label{fig:paramhop}
\end{figure*}

\begin{figure*}[t]
\centering
\includegraphics[width=17cm]{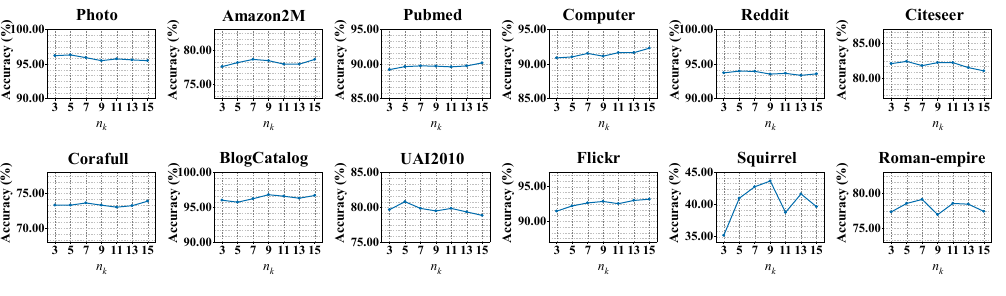}
\caption{
Performance of \name with different $n_k$ on all datasets. 
}
\label{fig:paramk}
\end{figure*}

\textbf{Analysis on Hops $K$.}
To validate the influence of the maximum range of neighborhood,
we fix the value of $n_k$ and vary $K$ in $\{3,5,\cdots, 15\}$.
\autoref{fig:paramhop} shows the results with different values of $K$.
We observe that the trends vary across different levels of homophily ratio. 
Specifically, the accuracy increases as $K$ increases on homophily graphs, while it exhibits the opposite trend on heterophily graphs. 
This is because that nodes are more likely to connect with similar nodes on homophily graphs.
In this case, during the neighborhood aggregation operation, as $K$ increases, nodes can gather information from more similar nodes, resulting in higher accuracy.
Meanwhile, nodes tend to connect with irrelevant nodes on heterophily graphs where increasing $K$ introduces more noise information, consequently hurting the model performance.

\textbf{Analysis of $n_k$.}
Similarly, we fix the value of $K$ and vary the number of selected nodes $n_k$ in $\{3,5,\cdots, 15\}$. 
The results are shown in \autoref{fig:paramk}. 
We can observe that \name exhibits low sensitivity to $n_k$ on most datasets.
Generally speaking, a small value of $n_k$ (\ie $<$ 11) can enable \name to achieve competitive results, while a large value of $n_k$ can affect the model performance.
This is because \name selects node with the highest similarity scores in both topology and attribute views as node tokens.
In this way, a small number of node tokens is sufficient to represent the various properties of the target node within the graph. 
While, a large value of $n_k$ can lead to a long node token sequence which could involve more irrelevant nodes, further hurting the model performance.
In addition, different graphs require different values of $n_k$ to achieve the best performance.
This is because that different types of graphs exhibit different structure and attribute features, which can affect the optimal value of $n_k$.

\section{Conclusion}
In this paper, we propose a novel graph Transformer called \name for the node classification task.
\name introduces a new token generator, \tname, that constructs two types of token sequences, node-wise and neighborhood-wise token sequences, from both topology and attribute views.
In this way, the graph information of nodes can be effectively extracted.
\name further incorporates a Transformer-based backbone with the adaptive fusion module to learn final node representations from the generated token sequences. 
Benefiting from the valuable graph information carried by the output of \tname, \name eliminates the need for graph-specific designs and solely relies on standard Transformer layers. 
This design enables \name to effectively learn distinct node representations across diverse graphs.
Extensive experimental results conducted on both homophily and heterophily graphs from small to large scale demonstrate the superiority of \name in the node classification task. 
In comparison to representative graph Transformers and GNNs, \name consistently achieves superior performance, affirming its effectiveness and potential in graph representation learning.

\section*{Acknowledgments}
This work is supported by National Natural Science Foundation (U22B2017).

%

\bibliographystyle{IEEEtran}
\bibliography{reference}

\vfill

\vfill

\end{document}